\documentclass[11pt,a4paper]{article}
\usepackage[hyperref]{emnlp-ijcnlp-2019}
\usepackage{times}
\usepackage{latexsym}
\usepackage{enumitem}
\usepackage{url}
\usepackage{balance}

\usepackage{graphicx}
\usepackage{amsmath}
\usepackage{subcaption}
\usepackage{amsfonts}
\usepackage{tabulary,booktabs}
\usepackage{multirow}

\newcolumntype{L}[1]{>{\raggedright\let\newline\\\arraybackslash\hspace{0pt}}m{#1}}
\newcolumntype{C}[1]{>{\centering\let\newline\\\arraybackslash\hspace{0pt}}m{#1}}
\newcolumntype{R}[1]{>{\raggedleft\let\newline\\\arraybackslash\hspace{0pt}}m{#1}}
\newcommand\Tstrut{\rule{0pt}{2.0ex}}         % = `top' strut
\aclfinalcopy % Uncomment this line for the final submission

\newcommand{\modelacronym}{GSAMN}

\title{A Gated Self-attention Memory Network for Answer Selection}

\author{Tuan Lai \thanks{\; Equal contributions. The work was conducted while the first author interned at Adobe Research.} \,\textsuperscript{1}, Quan Hung Tran \footnotemark[1] \,\textsuperscript{2},  \textbf{Trung Bui \textsuperscript{2}},  \textbf{Daisuke Kihara \textsuperscript{1}} \\
        \{lai123,dkihara\}@purdue.edu, \{qtran,bui\}@adobe.com
        \\
	    \textsuperscript{1} Purdue University, West Lafayette, IN\\
	    \textsuperscript{2} Adobe Research, San Jose, CA\\
}

\date{}

\begin{document}

\maketitle
\begin{abstract}
Answer selection is an important research problem, with applications in many areas. Previous deep learning based approaches for the task mainly adopt the Compare-Aggregate architecture that performs word-level comparison followed by aggregation. In this work, we take a departure from the popular Compare-Aggregate architecture, and instead, propose a new gated self-attention memory network for the task. Combined with a simple transfer learning technique from a large-scale online corpus, our model outperforms previous methods by a large margin, achieving new state-of-the-art results on two standard answer selection datasets: TrecQA and WikiQA.
\end{abstract}

\section{Introduction and Related Work}
Answer selection is an important task, with applications in many areas \cite{lai-etal-2018-review}. Given a question and a set of candidate answers, the task is to identify the most relevant candidate. Previous work on answer selection typically relies on feature engineering, linguistic tools, or external
resources \cite{trec_qa_dataset,Wang:2010:PTM:1873781.1873912,Heilman:2010:TEM:1857999.1858143,question-answering-using-enhanced-lexical-semantic-models,Yao13answerextraction}. Recently, with the renaissance of neural network models, many deep learning based methods have been proposed to address the task \cite{Tay2017LearningTR,Shen2017InterWeightedAN,Wang2017BilateralMM,Bian2017ACM,tymoshenko-moschitti-2018-cross,Tay2018MultiCastAN,tayyar-madabushi-etal-2018-integrating,Yoon2019ACM}. They outperform traditional techniques. A common trait of a number of these deep learning methods is the use of the Compare-Aggregate architecture~\cite{Wang2017ACM}. Typically in this architecture, contextualized vector representations of small units such as words of the question and the candidate are first \textit{compared and aligned}. After that, these comparison results are then \textit{aggregated} to calculate a score indicating the relevance between the question and the candidate. On standard answer selection datasets such as TrecQA \cite{trec_qa_dataset} or WikiQA \cite{Yang2015WikiQAAC}, Compare-Aggregate approaches achieve very competitive performance. However, they still have some limitations. For example, the first few layers of most previous Compare-Aggregate models encode the question-candidate pair into sequences of contextualized
vector representations separately \cite{Wang2017BilateralMM,Shen2017InterWeightedAN,Bian2017ACM}. These sequences are independent and completely ignore the information from the other sequence.

In this work, we take a departure from the popular Compare-Aggregate architecture,
so instead, we propose a mix between two very successful 
architectures in machine comprehension and sequence modeling,
the memory network~\cite{sukhbaatar2015end} and the 
self-attention architecture~\cite{vaswani2017attention}. In the context of answer selection, the self-attention architecture allows us to learn the contextual representation of elements in
the sequence with respect to both the question and the answer, while the multi-hop reasoning memory network allows us to refine the decision over multiple steps. To this end, we propose a new memory-based, gated self-attention architecture for the task of answer selection. Combined with a simple transfer learning technique from a large-scale online corpus, our model achieves new state-of-the-art results on the TrecQA and WikiQA datasets.

In the following parts, we first describe our gated self-attention memory network for answer selection in Section 2. We then go into details our transfer learning approach in Section 3. After that, we describe the conducted experiments and their results in Section 4. Finally, we conclude this work in Section 5.

\section{Gated Self-Attention Memory Network}
\subsection{The gated self-attention mechanism}

The gated attention mechanism~\cite{dhingra2016gated,tran2017generative} extends the popular scalar-based attention
mechanism by calculating a real vector gate to control the flow of information, instead of a scalar value. Let's denote the sequence of
input vectors as $X = [\textbf{x}_1..\textbf{x}_n]$. If we have context information $\textbf{c}$,
then in traditional attention mechanism, association score $\alpha_i$  is usually calculated as a normalized dot product between the two vectors $\textbf{c}$ and $\textbf{x}_i$ (Equation~\ref{eqn:attention_traditional}) where $i\in [1..n]$.
\begin{equation}
\begin{split}
    \alpha_i & = \frac{\exp(\textbf{c}^T\,\textbf{x}_i)}{\sum_{j \in [1..n]}\exp(\textbf{c}^T\,\textbf{x}_j)}
\end{split}
\label{eqn:attention_traditional}
\end{equation}

For the gated attention mechanism, the association between two vectors $\textbf{c}$ and $\textbf{x}_i$
is represented by gate vector $\textbf{g}_i$ as follows:
\begin{equation}
    \textbf{g}_i = \sigma\big(f(\textbf{c},\textbf{x}_i)\big) 
\label{eqn:attention_gate}
\end{equation}
where $\sigma$ denotes the element-wise sigmoid function. Function $f$ is a parameterized function and thus, is more flexible in modelling the interaction between vectors $\textbf{c}$ and $\textbf{x}_i$.

In this work, we propose a new type of self-attention based on the gated attention mechanism described above, and we refer to it as the \textit{gated self-attention} mechanism (GSAM). We want to condition the gate vector not only on a context vector and a single input vector but also on the entire sequence of inputs. Therefore, we design function $f$ to be dependent on all the inputs in the sequence and the context vector. To calculate the gate for input $\textbf{x}_i$, first, each of the inputs in the input sequence and the context vector will present an individual gate ``vote''. Then, the votes are aggregated to calculate gate $\textbf{g}_i$ for $\textbf{x}_i$. This process is illustrated in Equation~\ref{eqn:attention_gate_seq}:
%For a more compact annotation, in 
%Equation~\ref{eqn:attention_gate_seq}, we treat the context vector $\textbf{c}$ as
%another input $\textbf{x}_{n+1}$.
\begin{equation}
\begin{split}
    \textbf{v}^j &= \textbf{W} \textbf{x}_j + \textbf{b} \; ; \; \textbf{v}^c = \textbf{W} \textbf{c}  + \textbf{b}       \\
    % \textbf{q}^j &= \textbf{W}_q\,\textbf{x}_j \; ; \; \textbf{q}^c = \textbf{W}_q\,\textbf{c}         \\
    s^j_i & = \textbf{x}^T_i \textbf{v}^j \; ; \; s^c_i = \textbf{x}^T_i \textbf{v}^c \\
    \alpha^j_i & = \frac{\exp(s^j_i)}{\sum_{k \in [1..n]}\exp(s^k_i) + \exp(s^c_i)} \\
    \alpha^c_i & = \frac{\exp(s^c_i)}{\sum_{k \in [1..n]}\exp(s^k_i) + \exp(s^c_i)} \\
    \textbf{g}_i & = f_i(c,X) \\
    &= \sigma\Bigg(\sum_j\bigg(\alpha^j_i\,\textbf{x}^j \bigg) + \alpha^c_i\,\textbf{c}\Bigg)
\end{split}
\label{eqn:attention_gate_seq}
\end{equation}
where $\textbf{W}$ and $\textbf{b}$ are learnable parameters shared among
functions $f_1\,...\,f_n$. Vectors $\textbf{v}$s are linear-transformed 
inputs which are used to calculate the self attentions. 
$s^j_i$ is
the unnormalized attention score of input $\textbf{x}_j$ put on $\textbf{x}_i$ and $\alpha^j_i$ 
is the normalized score. We use affine-transformed inputs $\textbf{v}$ and $\textbf{x}$ to calculate 
the self-attention 
instead
of just $\textbf{x}$ to break the attention symmetry phenomenon.
%, which means when an
% input element $\textbf{x}_i$ votes to put a high attention to $\textbf{x}_j$, $\textbf{x}_j$ also attends strongly to $\textbf{x}_i$. This constraint does
% not hold true in all circumstances}

\subsection{Combining with the memory network}
\begin{figure}
    \centering
    \includegraphics[width=\linewidth]{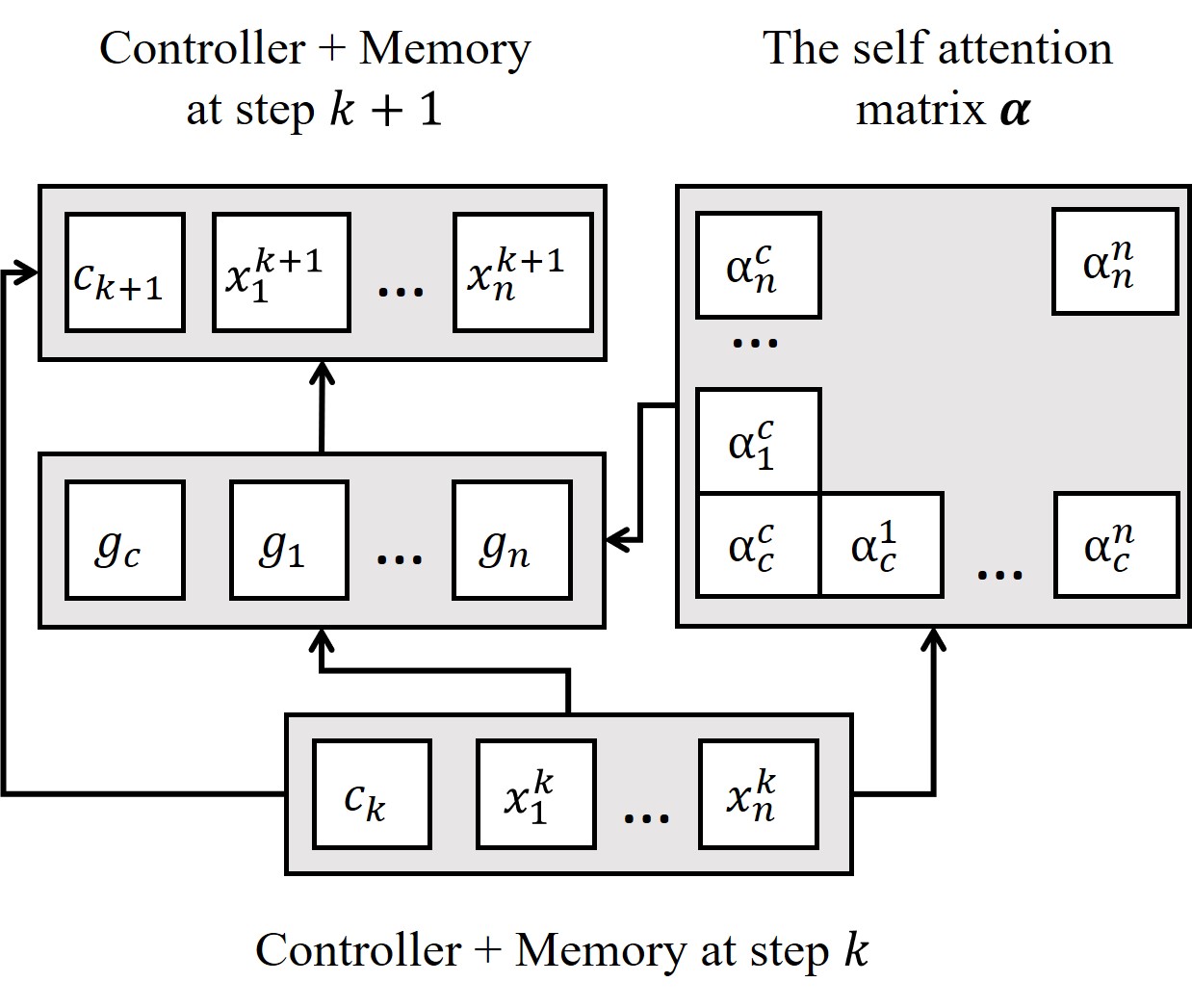}
    \caption{Simplified computation flow of the Gated Self-Attention Memory Network}
    \label{fig:architecture}
\end{figure}

In most previous memory network architectures, interactions between memory cells are relatively limited. At each hop, a single control vector is used to interpret each memory cell independently. To overcome this limitation, we combine GSAM described in Section 2.1 with the memory network architecture to create a new network called the Self-Attention Memory Network (GSAMN). Figure~\ref{fig:architecture} shows the simplified computation flow of GSAMN. In each reasoning hop, instead of using only context vector $\textbf{c}$
to interpret the inputs, we use GSAM. Let $\textbf{c}_k$ be the controlling context and $\textbf{x}_1^k\,...\,\textbf{x}_n^k$ be the memory values at the $k^\text{th}$ reasoning hop. Each memory cell update from the $k^\text{th}$ hop to the next hop is
%is a combination of the information from the updated controller and
calculated as the gated self-attention update (Equation~\ref{eqn:memory_gate_seq_mem}). 
\begin{equation}
\begin{split}
    \textbf{g}_i & = f_i(\textbf{c}_k,X) \\
    \textbf{x}^{k+1}_i & = \textbf{g}_i \odot \textbf{x}^{k}_i %+ (1-\textbf{g}_i)  \textbf{c}_{k+1} \\
\end{split}
\label{eqn:memory_gate_seq_mem}
\end{equation}

The controller's update is a combination of the gated self-attention above, and the memory network's traditional aggregate update. As the memory state values have already been attended to by the 
gating mechanism, we only need to average them (not weighted average).

\begin{equation}. 
\begin{split}
    \textbf{g}_c & = f_c(\textbf{c}_k,X) \\
    \textbf{c}_{k+1} & = \textbf{g}_c \odot \textbf{c}_{k} + \frac{1}{n}\sum_{i}\textbf{x}_i^{k+1}
\end{split}
\label{eqn:memory_gate_seq}
\end{equation}

\subsection{GSAMN for answer selection}
In the context of answer selection, we concatenate question Q and candidate answer A to a single sequence and treat the task as a binary classification problem. Given the GSAMN architecture above, we can use final controller
state $\textbf{c}_T$ as the representation of the sequence. The matching probability $P(A\,|\,Q)$ is finally calculated as follows:
\begin{equation}
\begin{split}
    P(A\,|\,Q) = \sigma\bigg(\textbf{W}_c\,\textbf{c}_T + \textbf{b}_c\bigg)
\end{split}
\label{eqn:final}
\end{equation}
where $\textbf{W}_c$ and $\textbf{b}_c$ are learnable parameters. We can initialize the memory values $\textbf{x}_1^0\,...\,\textbf{x}_n^0$ using any representation model such as word2vec \cite{Mikolov_word2vec}, GloVe \cite{Pennington2014GloveGV}, ELMo \cite{Peters2018DeepCW}, or BERT \cite{Devlin2018BERTPO}. The control vector $\textbf{c}$ is a randomly initialized learnable vector.

\section{Transfer Learning}
Previous studies on answer selection have focused mostly on small-scale datasets. On the other hand, many community question answering (CQA) platforms such as Yahoo Answers and Stack Exchange have become an essential source of information for many people. The amount of data (i.e., questions and answers) in these CQA platforms is huge and encompasses many domains and topics. This provides a great opportunity to apply transfer learning techniques to improve answer selection systems trained on limited datasets.

We crawled question-answer pairs related to various topics from Stack Exchange \footnote{\url{https://stackexchange.com/}}. After that, we removed every pair that contains text written in a language different from English. Furthermore, to ensure that in each collected pair the answer is highly relevant to the question, we removed pairs whose answers have less than two up-votes from community users. Finally, because training answer selection models also require negative examples, for each question, we sampled several real answers not related to the question to build up negative pairs. In the end, our dataset has 628,706 positive pairs and 9,874,758 negative pairs in total. We refer to our newly collected dataset as StackExchangeQA. Table \ref{datasets-examples} shows some examples of positive question-answer pairs from the dataset. The dataset has question-answer pairs from many different domains and topics. The code for constructing the StackExchangeQA dataset is available online \footnote{\url{https://github.com/laituan245/StackExchangeQA}}.

In this work, we employ a basic transfer learning technique. The first step is to pre-train our answer selection model on the StackExchangeQA dataset. Then, the second step is to fine-tune the same model on a target dataset of interest such as TrecQA or WikiQA. Despite the simplicity of the technique, the performance of our model improves substantially compared to not using transfer learning. Different from previous works which use source datasets that were manually annotated \cite{Min2017QuestionAT,Chung2018SupervisedAU}, our source dataset required minimal effort to obtain and preprocess. The choice of crawling question-answer pairs from the Stack Exchange website was arbitrary. We could also have crawled data from websites such as Yahoo Answers instead.

{\renewcommand{\arraystretch}{1.5}
\begin{table*}[!ht]
 \small
\begin{center}
\begin{tabular}{|l|p{125mm}|}
\hline Domain & QA Pair \\ \hline
Academia & \textbf{Question}: Is it okay for a PhD student to go on holidays in breaks? \newline \textbf{Answer}: Do you have an adviser? Have you talked to them about this? Most should be fine with you taking some time off to visit your family, but you should probably discuss longer breaks with them to work out all the details. \\ \hline
Apple Product & \textbf{Question}: What hidden features have you found in iOS 6? \newline \textbf{Answer}: Newly downloaded apps have a ``new'' label on the home screen.\\ \hline 
Chemistry & \textbf{Question}: Hydrochloric acid vs hydrogen chloride? \newline \textbf{Answer}: Hydrochloric acid is an aqueous solution of hydrogen chloride.\\ \hline
Cooking & \textbf{Question}: How do I prevent tomatoes from falling in a green salad? \newline \textbf{Answer}: I work around this by serving tomatoes on the top of the individual salads after they've been portioned out.  I'm not sure of a way to keep them incorporated.\\ \hline
Philosophy & \textbf{Question}: What did Socrates teach which lead to his conviction that he spoiled youth and taught other Gods? \newline \textbf{Answer}: I think in general one of the problems Socrates' contemporaries may have had with him was not so much what he taught but how he taught. Perhaps Socrates' method of philosophy was characterised more by testing propositions through questioning, than any strict concern with formulating a set of propositions on any one subject.\\ \hline
% Programming &  \textbf{Question}: C\# library for parsing HTML? \newline \textbf{Answer}: AngleSharp: (1) Actively developed/maintained (2) Built-in support for CSS selectors.\\ \hline
\end{tabular}
\end{center}
\caption{\label{datasets-examples} Examples of positive question-answer pairs from the StackExchangeQA dataset}
\end{table*}

\section{Experiments and Results}
To evaluate the effectiveness of our proposed answer selection model, we use two datasets: TrecQA and WikiQA. The TrecQA dataset \cite{trec_qa_dataset} was created from the TREC Question Answering tracks. There are two versions of TrecQA: raw and clean. Both versions have the same training set but their development and test sets differ. In this study, we use the clean version of the dataset that removed questions in development and test sets with no answers or only positive/negative answers. The clean version has 1,229/65/68 questions and 53,417/1,117/1,442 question-answer pairs for the train/dev/test split. The WikiQA dataset \cite{Yang2015WikiQAAC} was constructed from real queries of Bing and Wikipedia. Following the literature \cite{Yang2015WikiQAAC,Bian2017ACM,Shen2017InterWeightedAN}, we removed all questions with no correct answers before training and evaluating answer selection models. The excluded WikiQA has 873/126/243 questions and 8,627/1,130/2,351 question-answer pairs for train/dev/test split.

Similar to previous work, we report the model performance as the mean average precision (MAP) and mean reciprocal rank (MRR) \footnote{\url{https://aclweb.org/aclwiki/Question_Answering_(State_of_the_art)}}. In all experiments, we use the base version of BERT \cite{Devlin2018BERTPO} to initialize the memory of our proposed architecture. We fine-tune the BERT embeddings during training. We set the number of reasoning hops to be 2. We use the Adam optimizer with a learning rate of 5e-5, $\beta_1 = 0.9$, $\beta_2 = 0.999$, L2 weight decay of $0.01$, learning rate warmup over the first 10 percent of the total number of training steps, and linear decay of the learning rate. We did hyper-parameter tuning on the development sets.

It is worth noting that, we have experimented with various values for the number of reasoning hops. We found that using 2 hops gives the best performance on the tested datasets while using larger number of hops decreases the performance slightly.  We attribute the diminishing returns in increasing the number of hops to the limited size of the TrecQA and WikiQA datasets. Many previous works related to memory networks also use small number of memory hops \cite{Weston2015MemoryN,sukhbaatar2015end,Miller2016KeyValueMN,Zhang2018PersonalizingDA}.

\subsection{Comparison with Previous Methods} 
Table \ref{tab:TrecQA_WikiQA_overall} summarizes the performances of our proposed models and compares them to the baselines on the TrecQA and WikiQA datasets. The full model [BERT + \modelacronym + Transfer Learning] outperforms the previous state-of-the-art methods by a large margin. Note that by simply fine-tuning the pre-trained BERT embeddings, one can easily achieve very competitive performance on both datasets. This is expected as BERT has been pre-trained on a massive amount of unlabeled data. However, our proposed techniques do add a significant amount of performance. The gain from using all of our proposed techniques is larger than the difference between fine-tuning BERT model compared to previous systems in the TrecQA dataset.

{\renewcommand{\arraystretch}{1.0}
\begin{table*}[!ht]
\centering
{
\begin{tabular}{L{0.85\columnwidth}|C{0.2\columnwidth}C{0.2\columnwidth}|C{0.2\columnwidth}C{0.2\columnwidth}}
\hline
\multirow{3}{*}{\textbf{Model}}   & \multicolumn{2}{c|}{\textbf{TrecQA}}\Tstrut  & \multicolumn{2}{c}{\textbf{WikiQA}} \\ 
\cline{2-5}              & \multicolumn{1}{c}{MAP}\Tstrut   & \multicolumn{1}{c|}{MRR}   & \multicolumn{1}{c}{MAP}   & \multicolumn{1}{c}{MRR} \\
\hline \hline
BERT + \modelacronym + Transfer & \textbf{0.914} & \textbf{0.957} & \textbf{0.857} & \textbf{0.872} \Tstrut  \\
BERT + Transformers + Transfer & 0.895 & 0.939 & 0.831 & 0.848\Tstrut \\
\hline
BERT + \modelacronym & 0.906 & 0.949 & 0.821 & 0.832 \Tstrut \\
BERT + Transformers & 0.886 & 0.926 & 0.813 & 0.828 \Tstrut \\
ELMo + Compare-Aggregate & 0.850 & 0.898 & 0.746 & 0.762 \Tstrut \\
\hline
BERT + Transfer & 0.902 & 0.949 & 0.832  & 0.849 \Tstrut \\
BERT & 0.877 & 0.922 & 0.810 & 0.827 \Tstrut \\
\hline
QC + PR + MP CNN \shortcite{tayyar-madabushi-etal-2018-integrating} & 0.865 & 0.904 & --- & --- \Tstrut \\
IWAN + sCARNN \shortcite{Tran2018TheCA} & 0.829 & 0.875 & 0.716 & 0.722 \Tstrut\\
IWAN \shortcite{Shen2017InterWeightedAN} & 0.822 & 0.889 & 0.733 & 0.750 \Tstrut \\
Compare-Aggregate \shortcite{Bian2017ACM} & 0.821 & 0.899 & 0.748 & 0.758 \Tstrut\\
BiMPM \shortcite{Wang2017BilateralMM} & 0.802 & 0.875 & 0.718 & 0.731 \Tstrut \\
HyperQA \shortcite{HyperQA} & 0.784 & 0.865 & 0.705 & 0.720 \Tstrut \\
NCE-CNN \shortcite{rao2016noise} & 0.801 & 0.877 & --- & --- \Tstrut\\
Attentive Pooling CNN \shortcite{Santos2016AttentivePN} & 0.753 & 0.851 & 0.689 & 0.696 \Tstrut\\
W\&I \shortcite{Wang2015FAQbasedQA} & 0.746 & 0.820 & --- & --- \Tstrut\\

\end{tabular}
}
\caption{Results on the TrecQA and WikiQA datasets}
\label{tab:TrecQA_WikiQA_overall}
\end{table*}

\subsection{Ablation Analysis}
We aim to analyze the relative effectiveness of different components of our full model. From the original BERT baseline, we add one component at a time and evaluate the performance of the partial models on the datasets. From Table \ref{tab:TrecQA_WikiQA_overall}, we can see that both the variants [BERT + \modelacronym] and [BERT + Transfer Learning] have better performance than the original BERT baseline. However, both of the partial variants still perform worse than the one with all the techniques. This shows that although each of our proposed components is effective by itself,  we need to combine them together in order to achieve the best performance.

\subsection{GSAMN versus Transformer}
It was an iterative process to arrive at the current design of \modelacronym. We aim to analyze whether the improvement in performance comes from the inductive bias that we introduced into the architecture or simply from having more parameters due to added complexity. To this end, we evaluated the performances of two variants [BERT + Transformers] and [BERT + Transformers + Transfer Learning]. These model variants simply use two Transformer layers \cite{vaswani2017attention} on top of BERT instead of using our \modelacronym\, architecture. 
Table \ref{tab:TrecQA_WikiQA_overall} clearly shows that \modelacronym\ 
outperforms the Transformer based variants, with or without the transfer 
learning component.

We have experimented with adding more Transformer layers on top of BERT but the performance did not improve. For example, using 6 extra Transformer layers only achieves a MAP score of 0.885 on the TrecQA dataset. This is reasonable because BERT by itself already contains 12 Transformer layers. Without a new kind of layer such as our proposed GSAMN architecture, stacking more Transformer layers will not be helpful, especially in this case where the tested datasets are not large.

\subsection{GSAMN versus Compare-Aggregate}
Finally, we have a comparison between our full model and the Compare-Aggregate framework. Most previous Compare-Aggregate architectures use traditional word embeddings such as word2vec \cite{Mikolov_word2vec} or GloVe \cite{Pennington2014GloveGV}. In contrast, our full model uses BERT which is an arguably more powerful language representation model. To this end, we implemented a Compare-Aggregate variant that uses dynamic-clip attention \cite{Bian2017ACM}. We use ELMo \cite{Peters2018DeepCW} to represent the input words to the implemented Compare-Aggregate architecture. We use ELMo instead of BERT because BERT is in subword-level while one of the intuitions behind the Compare-Aggregate variant is about comparing word-level representations. In addition, we have tested the variant [BERT + Compare-Aggregate] but found it to be worse than the version [ELMo + Compare-Aggregate]. The results in Table \ref{tab:TrecQA_WikiQA_overall} show that our model significantly outperforms [ELMo + Compare-Aggregate] as well.

\section{Conclusions}
In this paper, we propose a new gated self-attention memory network architecture 
for answer selection. Combined with a simple transfer learning 
technique from a large-scale CQA corpus, the model achieves the state-of-the-art performance on two well-studied answer selection 
datasets: TrecQA and WikiQA.  
 In the future, we plan to investigate more transfer learning techniques for utilizing the large volume of existing CQA data. In addition, we plan to apply our self-attention memory network on other sentence matching tasks such as natural language inference, paraphrase identification, or measuring semantic relatedness.

\balance

\bibliography{emnlp-ijcnlp-2019}
\bibliographystyle{acl_natbib}

% \appendix
% \section{Datasets}
% \input{appendix_dataset.tex}

\end{document}